\documentclass[letterpaper, 10pt]{article}
\usepackage{graphicx}
\usepackage{amsmath}
\usepackage{float}
\usepackage{amsfonts}
\usepackage[inner=0.75in, outer=0.75in, top=0.75in, bottom=1in, paperheight=11in, paperwidth=8.5in]{geometry}

\usepackage{epstopdf}
\usepackage{float}
\usepackage{amssymb}
\usepackage[utf8]{inputenc}
\usepackage[english]{babel}
\usepackage{amsthm}
\usepackage{color}
\usepackage{forest}

\usepackage{microtype}
\usepackage{url}
\usepackage{subfiles}
\usepackage{caption}
\usepackage{subcaption}
\usepackage{parskip}
\usepackage[section]{placeins}

\usepackage{xcolor}
\usepackage{framed}

\colorlet{shadecolor}{blue!20}

\usepackage[obeyFinal,textwidth=0.55in]{todonotes}
\usepackage[hidelinks]{hyperref}
\usepackage[capitalize]{cleveref}

\theoremstyle{definition}

\begin{document}	

\title{\bf On partitioning of an SHM problem and parallels with transfer learning}	
\author{G.\ Tsialiamanis, D.J.\ Wagg, P.A.\ Gardner, N.\ Dervilis \& K.\ Worden\\
        $^1$Dynamics Research Group, Department of Mechanical Engineering, University of Sheffield \\
        Mappin Street, Sheffield S1 3JD, UK\\}
	   
	\date{}
    \maketitle
	\thispagestyle{empty}

\section*{Abstract}
In the current work, a problem-splitting approach and a scheme motivated by transfer learning is applied to a structural health monitoring problem. The specific
problem in this case is that of localising damage on an aircraft wing. The original experiment is described, together with the initial approach, in which a neural 
network was trained to localise damage. The results were not ideal, partly because of a scarcity of training data, and partly because of the difficulty in 
resolving two of the damage cases. In the current paper, the problem is split into two sub-problems and an increase in classification accuracy is obtained. The 
sub-problems are obtained by separating out the most difficult-to-classify damage cases. A second approach to the problem is considered by adopting ideas from 
transfer learning (usually applied in much deeper) networks to see if a network trained on the simpler damage cases can help with feature extraction in the more
difficult cases. The transfer of a fixed trained batch of layers between the networks is found to improve classification by making the classes more separable in the feature
space and to speed up convergence.

\textbf{Key words: Structural health monitoring (SHM), machine learning, classification, problem splitting, transfer learning.}

\section{Introduction}
\label{sec:intro}

Structural health monitoring (SHM) refers to the process of implementing a damage detection strategy for aerospace, civil or mechanical engineering infrastructure \cite{farrar2012structural}. Here, damage is defined as changes introduced into a system/structure, either intentionally or unintentionally, that affect current or 
future performance of the system. Detecting damage is becoming more and more important in modern societies, where everyday activities depend increasingly on 
engineering systems and structures. One the one hand, safety has to be assured, both for users and for equipment or machinery existing within these structures. On 
the other hand, infrastructure is often designed for a predefined lifetime and damage occurrence may reduce the expected lifetime and have a huge economic impact 
as a result of necessary repairs or even rebuilding or decommison. Damage can be visible on or in structures, but more often it is not, and has to be inferred from 
signals measured by sensors placed on them.

An increasingly useful tool in SHM is {\em machine learning} (ML) \cite{farrar2012structural}. In many current applications large sets of data are gathered by sensors 
or generated by models and these can be exploited to gain insight into structural dynamics and materials engineering. Machine learning is employed because of its 
efficiency in classification, function interpolation and prediction using data. Data-driven models are built and used to serve SHM purposes. These models can also be 
used to further understand how structures react to different conditions and explain their physics. However, one of the main drawbacks of such methods is the need for 
large datasets. ML models may have many parameters which are established during {\em training} on data which may need to span all the health conditions of interest for the
given structure or system. Larger datasets assist in better tuning of the models as far as accuracy and generalisation are concerned. However, even if large datasets are available, sometimes there are very few observations on damaged states, which are important in SHM. In the current paper, increased 
accuracy of a data-driven SHM classifier will be discussed in terms of two strategies: splitting the problem into two sub-problems and attempting transfer of information
between the two sub-problems in a manner motivated by transfer learning \cite{Pan2010}.

\begin{figure}[ht!]
	\centering
	\begin{subfigure}{.5\textwidth}
		\centering
		\includegraphics[width=.95\linewidth]{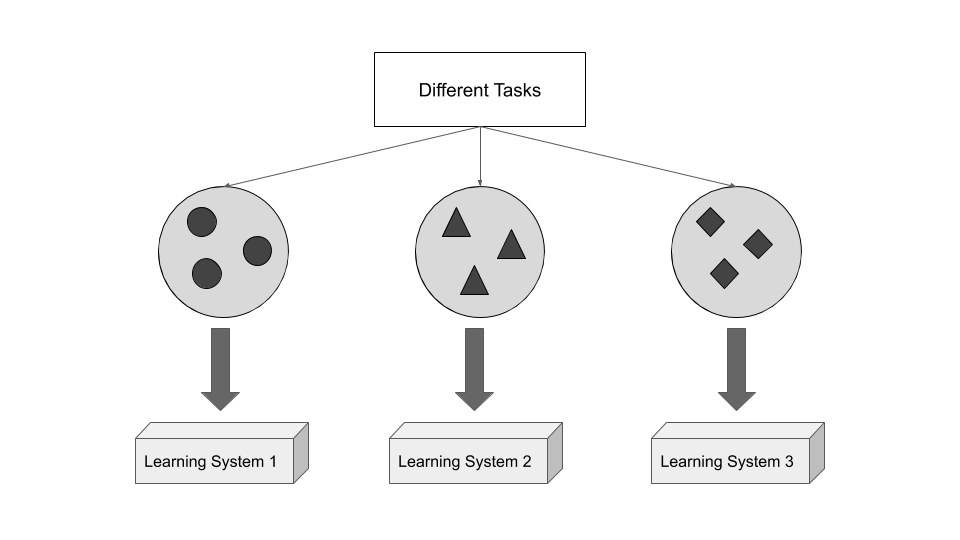}
		\caption{}
		\label{fig:trad_ml}
	\end{subfigure}%
	\begin{subfigure}{.5\textwidth}
		\centering
		\includegraphics[width=.95\linewidth]{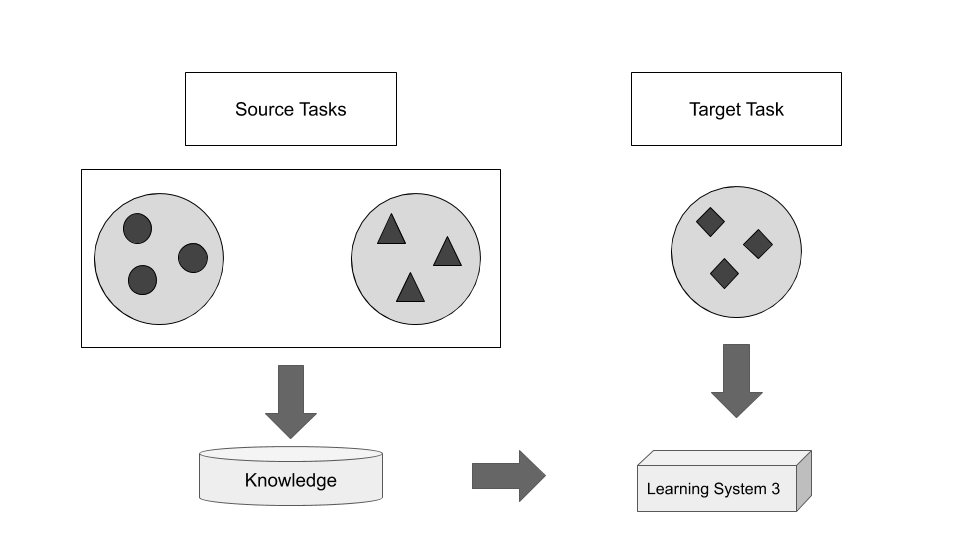}
		\caption{}
		\label{fig:transfer_ml}
	\end{subfigure}\\
	\caption{Traditional (a) and transfer (b) learning schemes (following \cite{Pan2010}).}
	\label{fig:ml_schemes}
\end{figure}

Transfer learning is the procedure of taking knowledge from a source domain and task and applying it to a different domain and task to help improve performance on the 
second task \cite{Pan2010}. Transfer learning is useful, as we know that a model trained on a dataset can not naturally be applied on another due to difference in data distribution, but can be further tuned to also apply on the second dataset. An accurate representation of the difference between traditional and transfer learning schemes can be seen in Figure \ref{fig:ml_schemes}. 
The SHM problem herein will be addressed using neural networks \cite{Bishop:1995:NNP:525960}, for which transfer learning has been proven quite efficient (although 
usually in deeper learning architectures \cite{GabrielPuiCheongFung2006,AlMubaid2006}). Due to the layered structure of the networks, after having created a model for a task, transferring a part of it (e.g.\ some 
subset of the layers) is easy. The method is used in many disciplines, such as computer vision \cite{oquab2014learning,shin2016deep}. The most commonly-used learners
are Convolutional Neural Networks (CNNs), which can be very slow to train and may need a lot of data, which in many cases can be hard to obtain (e.g.\ labelled 
images). These problems can be dealt with by using the fixed initial layers of pre-trained models to extract features of images, and then train only the last layers to 
classify in the new context. In this way, both the number of trainable parameters and the need for huge datasets and computation time are reduced. Another topic that 
transfer learning has been used in is natural language processing (NLP) \cite{DBLP:journals/corr/BingelS17}, where the same issues of lack of labelled data and large 
amounts of training time are dealt with by transferring of pre-trained models into new tasks. Further examples of the benefits of transfer learning can be found in web 
document classification \cite{GabrielPuiCheongFung2006,AlMubaid2006}; in these cases, in newly-created web sites, lack of labelled data occurs. To address this problem, 
even though the new web sites belong to a different domain than the training domain of the existing sites, the same models can be used to help classify documents in the 
new websites.

In the context of the current work, transfer learning is considered in transferring knowledge from one sub-problem to the other by introducing pre-trained layers into
new classifiers. The classification problem that will be presented is related to damage class/location. A model trained to predict a subset of the damage classes (source 
task) with data corresponding of that subset (source domain), will be used to boost performance of a second classifier trained to identify a different subset of damage
states. 

\section{Problem description}
\label{sec:problem_desc}

Similar to the aforementioned applications, in SHM machine learning is also used for classification and regression. In data driven SHM one tries to identify features that will reveal whether a structure is damaged or what type of damage is present and so, labelled data are necessity. Therefore, in SHM applications lack of labelled data about damage location or severity is a drawback. SHM problems can be categorised in many ways but are often broken down according to the hierarchical structure proposed by Rytter \cite{rytter1993vibrational}:  

\begin{enumerate}
	\item Is there damage in the system ({\em existence})?
	\item Where is the damage in the system ({\em location})?
	\item What kind of damage is present ({\em type/classification})?
	\item How severe is the damage ({\em extent/severity})?
	\item How much useful (safe) life remains ({\em prognosis})?
\end{enumerate}

A common approach to the first level is to observe the structure in its normal condition and try to find changes in features extracted from measured signals that 
are sensitive to damage. This approach is called {\em novelty detection} \cite{Worden1997,WORDEN2000}, and it has some advantages and disadvantages. The main advantage 
is that it is usually an {\em unsupervised} method, that is only trained on data that are considered to be from the undamaged condition of the structure, without 
a specific target class label. These methods are thus trained to detect any {\em changes} in the behaviour of the elements under consideration, which can be a 
disadvantage, since structures can change their behaviour for benign reasons, like changes in their environmental or operational conditions; such benign changes or
{\em confounding influences} can raise false alarms.

In this work a problem of damage localisation is considered (at Level 2 in Rytter's hierarchy \cite{rytter1993vibrational}); the structure of interest being a wing of 
a Gnat trainer aircraft. The problem is one of supervised-learning, as the data for all damage cases were collected and a classification model was trained accordingly. Subsequently, the classifier was used to predict the damage class of newly-presented data. The features used as inputs to the classifier were novelty indices calculated 
between frequency intervals of the transmissibilities of the normal condition of the structure (undamaged state) and the testing states. The transmissibility between 
two points of a structure is given by equation (\ref{eq:transmissibility}), and this represents the ratio of two response spectra. This feature is useful because it 
describes the response of the structure in the frequency domain, without requiring any knowledge of the frequency content of the excitation. The transmissibility is
defined as,

\begin{equation} 
\label{eq:transmissibility}
     T_{ij} = \frac{FRF_i}{FRF_j} =  
     \frac{\frac{\mathcal{F}_{i}}{\mathcal{F}_{excitation}}}{\frac{\mathcal{F}_{j}}{\mathcal{F}_{excitation}}} = \frac{\mathcal{F}_{i}}{\mathcal{F}_{j}}
\end{equation}
where, $\mathcal{F}_{i}$ is the Fourier Transform of the signal given by the $i^{th}$ sensor and $FRF_i$ is the {\em Frequency Response Function} (FRF) at the $i$th 
point. 

\begin{figure}[ht!]
	\centering
	\includegraphics[scale=0.60]{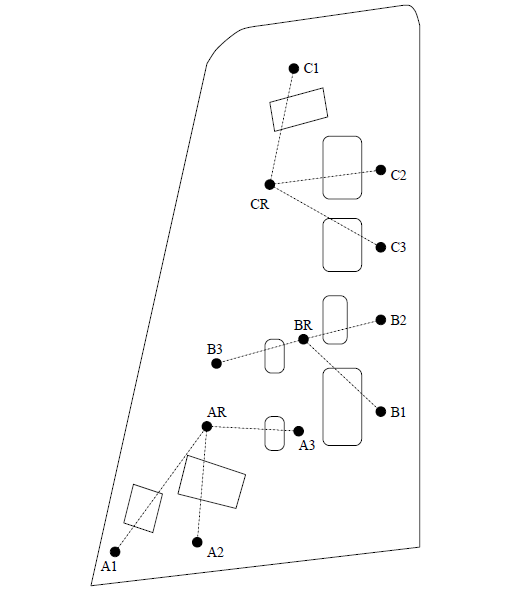}
	\caption{Configuration of sensors on the Gnat aircraft wing \cite{manson2003experimental}.}
	\label{fig:sensors}
\end{figure}

The experiment was set up as described in \cite{Worden2007}. The wing of the aircraft was excited with a Gaussian white noise using an electrodynamic shaker attached 
on the bottom surface of the wing. The configuration of the sensors placed on the wing can be seen in Figure \ref{fig:sensors}. Responses were measured with 
accelerometers on the upper surface of the wing, and the transmissibilities between each sensor and the corresponding reference sensor were calculated. The 
transmissibilities were recorded in the 1-2 kHz range, as this interval was found to be sensitive to the damage that was going to be introduced to the structure. Each transmissibility contained 2048 spectral lines.

\begin{figure}[h!]
	\centering
	\includegraphics[scale=0.70]{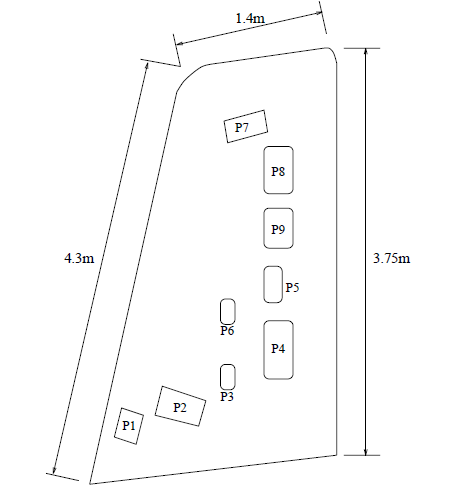}
	\caption{Schematic showing wing panels removed to simulate the nine damage cases \cite{manson2003experimental}.}
	\label{fig:panels}
\end{figure}

Initially, the structure was excited in its normal condition, i.e.\ with no introduced damage. The transmissibilities of this state were recorded and subsequently, 
to simulate damage, several panels were removed from the wing, one at a time. In each panel removal, the wing was excited again with white Gaussian noise and the transmissibilities were recorded. The panels that were removed are shown in Figure \ref{fig:panels}. Each panel has a different size, varying from 0.008 to 0.08m$^{2}$ 
and so the localisation of smaller panels becomes more difficult, since their removal affects the transmissibilities less than the bigger panels. The measurements were 
repeated 200 times for each damage case, ultimately leading to 1800 data points belonging to nine different damage cases/classes. The data were separated into training, 
validation and testing sub sets, each having 66 points per damage case.

For the purposes of damage localisation, features had to be selected which would be sensitive to the panel removals; this was initially done manually \cite{manson2003experimental}, selecting by visual `engineering judgement' the intervals of the transmissibilities that appeared to be more sensitive to damage and 
calculating the novelty indices of each state by comparison with the transmissibilities of the undamaged state. The novelty indices were computed using the 
Mahalanobis squared-distance (MSD) $D^2_{\zeta}$ of the feature vectors $\mathbf{x_{\zeta}}$, which in this case contained the magnitudes of transmissibility spectral 
lines. The MSD is defined by,

\begin{equation} 
\label{eq:mahal_dist}
     D_{\zeta}^{2} = (\mathbf{x_{\zeta}} - \mathbf{\overline{x}})^{T} S^{-1}(\mathbf{x_{\zeta}} - \mathbf{\overline{x}})
\end{equation}
were $\mathbf{\overline{x}}$ is the sample mean on the normal condition feature data, and $S$ is the sample covariance matrix.

After selecting `by eye' the most important features for damage detection \cite{manson2003experimental}, a genetic algorithm was used \cite{Worden2008} to choose the 
most sensitive features, in order to localise/classify the damage. Finally, nine features were chosen as the most sensitive and an MLP neural network \cite{Bishop:1995:NNP:525960} with nine nodes in the input layer, ten nodes in the hidden layer and nine nodes in the decision layer was trained. The confusion matrix 
of the resulting classifier is shown in the Table \ref{Tab:Initial_network_conf_mat}. It can be seen that the misclassification rate is very low and that the damage 
cases that are most confused are the ones where the missing panel is Panel 3 or Panel 6, which were the smallest ones.

\section{Problem splitting}
\label{sec:splitting}

As mentioned in \cite{tarassenko1998guide}, the rule-of-thumb for a network that generalises well is that it should be trained with at least ten samples per weight 
of the network. The aforementioned network had 180 trainable weights (and another 19 bias terms) so the 596 training samples are not ideal for the neural network. As a solution, 
a splitting of the original problem into two sub-problems is considered here to try and reduce the misclassification rate on the testing data even further. The dataset 
is split into two parts, one containing all the damage cases except Panels 3 \& 6 and the second containing the rest of the data. Subsequently, two neural network 
classifiers were trained separately on the new datasets. This was thought to be a good practice, since the panels are the smallest, and their removal affects the 
novelty indices less than the rest of the panel removals. The impact is that the points appear closer to each other in the feature space, and are swamped by points 
belonging to other classes, so the initial classifier cannot separate them efficiently. By assigning the tasks to different classifiers, an increase in performance 
is expected, especially in the case of separating the two smallest panel classes.

\begin{table}[h!]
	\centering
	\begin{tabular}{ |p{3cm}||p{1.0cm}|p{1.0cm}|p{1.0cm}|p{1.0cm}|p{1.0cm}|p{1.0cm}|p{1.0cm}|p{1.0cm}|p{1.0cm}| }
		\hline
		Predicted panel & 1 & 2 & 3& 4& 5 & 6 & 7 & 8 & 9\\
		\hline
		Missing panel 1& 65 & 0   & 0 & 0 & 0 & 0 & 0 & 0 & 1 \\
		Missing panel 2& 0 & 65  & 0 & 1 & 0 & 0 & 0 & 0 & 0 \\
		Missing panel 3& 1 &  0    & 62 & 0 & 0 & 1 & 0 & 1 & 1 \\
		Missing panel 4 & 0 &  0  & 0 & 66 & 0 & 0 & 0 & 0 & 0 \\
		Missing panel 5& 0 &  0  & 0 & 0 & 66 & 0 & 0 & 0 & 0 \\
		Missing panel 6& 0 &  3  & 0 & 0 & 0 & 62 & 0 & 1 & 0 \\
		Missing panel 7& 0 &  0  & 0 & 0 & 0 & 0 & 66 & 0 & 0 \\
		Missing panel 8& 1 &  0  & 0 & 0 & 0 & 0 & 0 & 65 & 0 \\
		Missing panel 9& 0 &  0  & 0 & 0 & 0 & 0 & 0 & 0 & 66 \\
		\hline
	\end{tabular}
	\caption{Confusion Matrix of neural network classifier, test set, total accuracy: 98.14\% \cite{Worden2008}}
	\label{Tab:Initial_network_conf_mat}
\end{table}

\begin{figure}[ht!]
	\centering
	\begin{subfigure}{.5\textwidth}
		\centering
		\includegraphics[width=.98\linewidth]{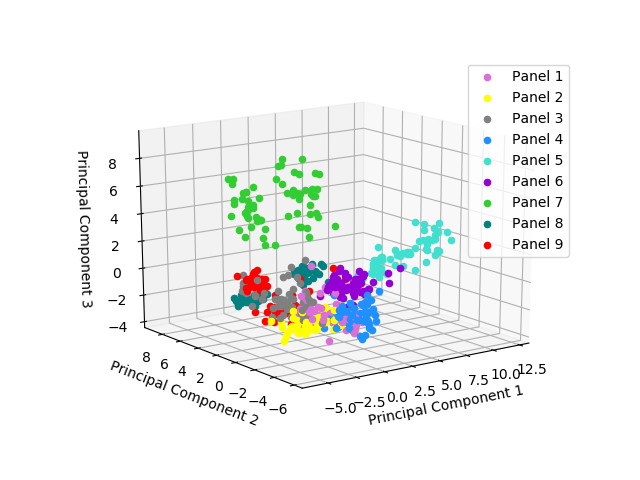}
		\caption{}
		\label{fig:all_data_pcs}
	\end{subfigure}%
	\begin{subfigure}{.5\textwidth}
		\centering
		\includegraphics[width=.98\linewidth]{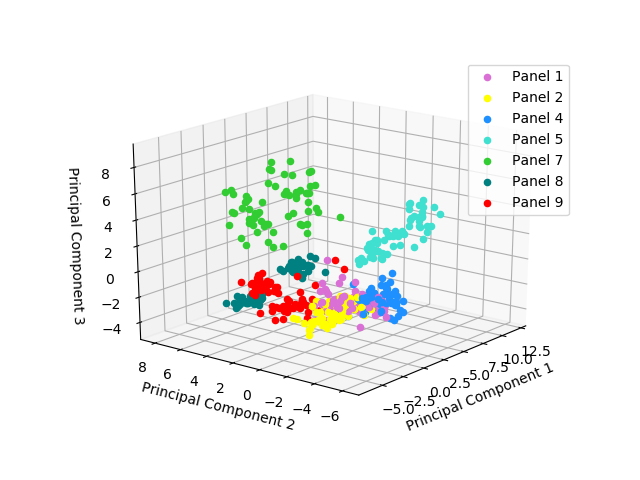}
		\caption{}
		\label{fig:7_classes_pcs}
	\end{subfigure}\\
	\begin{subfigure}{.5\textwidth}
		\centering
		\includegraphics[width=.98\linewidth]{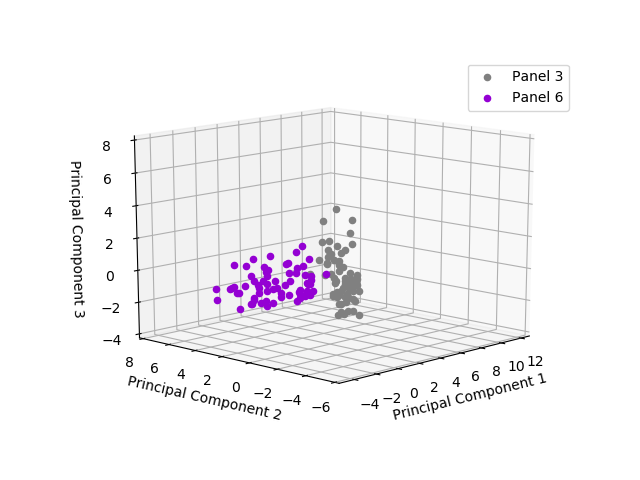}
		\caption{}
		\label{fig:2_classes_pcs}
	\end{subfigure}
	\caption{Principal components of all samples (a), samples excepting panels 3 and 6 (b) and samples of panels 3 and 6(c).}
	\label{fig:first_pcs}	
\end{figure}

To illustrate the data feature space, a visualisation is attempted here. Since the data belong to a nine-dimensional feature space, principal component analysis (PCA) 
was performed on the data and three of the principal components, explaining 71\% of total variance, are plotted in scatter plots shown in Figure \ref{fig:all_data_pcs}. 
Points referring to data corresponding to the missing panels 3 and 6 (grey and magenta points respectively) are entangled with other class points causing most of the misclassification rate shown above.

Random initialisation was followed for the neural networks. Initial values of the weights and biases of the networks were sampled from a normal zero-mean distribution. The two networks were initialised several times and trained for different sizes of the hidden layer to find the ones with optimal structure for the 
newly-defined problems. After randomly initialising and training multiple neural networks for both cases and keeping the ones with the minimum loss function value 
the best architectures were found to be networks with nine nodes in the hidden layer for both cases and seven output nodes for the first dataset and two for the 
second. The loss function used in training was the categorical cross-entropy function given by,

\begin{equation} 
\label{eq:categorical_crossentropy}
     L(y, \hat{y}) = -\frac{1}{N}\sum_{i=1}^{N}\sum_{j=1}^{n_{cl}} [y_{i, j}log \hat{y}_{i, j} + (1 - y_{i, j}) log (1 - \hat{y}_{i, j})]
\end{equation}

In Equation (\ref{eq:categorical_crossentropy}), $N$ is the number of samples during training, $n_{cl}$ is the number of possible classes, $\hat{y}_{i,j}$ the estimated probability that the $i$th point belongs to the $j$th class and $y_{i, j}$ is 1 if the $i$th sample belongs to the $j$th class, otherwise it is 0.

Confusion matrices on the test sets for the classifiers are shown in Tables \ref{Tab:First_set_conf_mat} and \ref{Tab:Second_set_conf_mat}. By splitting the dataset 
into two subsets the total accuracy is slightly increased from 98.14\% to 98.82\%. This is best considered in terms of classification error, which has been reduced from
1.86\% to 1.18\%, and this is an important reduction in SHM terms. Reduction of the number of trainable parameters has certainly contributed to this improvement, since 
the amount of training data is small. Performance on the task of separating only the two smallest panel classes was also increased because it is an easier task for the 
classifier than trying to discriminate them among the panel removals with greater impact on the novelty indices. This fact is also clear in Figure \ref{fig:2_classes_pcs}, 
where the principal components of samples belonging to the classes of missing Panels 3 and 6 are clearly separable.

\begin{table}[ht!]
	\centering
	\begin{tabular}{ |p{3cm}||p{1.0cm}|p{1.0cm}|p{1.0cm}|p{1.0cm}|p{1.0cm}|p{1.0cm}|p{1.0cm}| }
		\hline
		Predicted panel & 1 & 2 & 4& 5 & 7 & 8 & 9\\
		\hline
		Missing panel 1& 65 & 1   & 0 & 0 & 0 & 0 & 0 \\
		Missing panel 2& 0 & 63  & 1 & 0 & 0 & 0 &  2\\
		Missing panel 4 & 1 &  0  & 65 & 0 & 0 & 0 & 0 \\
		Missing panel 5& 0 &  0  & 0 & 66 & 0 & 0 & 0 \\
		Missing panel 7& 0 &  0  & 0 & 0 & 66 & 0 & 0 \\
		Missing panel 8& 1 &  0  & 0 & 0 & 0 & 65 & 0 \\
		Missing panel 9& 1 &  0  & 0 & 0 & 0 & 0 & 65 \\
		\hline
	\end{tabular}
	\caption{Confusion Matrix of neural network classifier trained on the first dataset, test set, total accuracy: 98.48\%}
	\label{Tab:First_set_conf_mat}
\end{table}

\begin{table}[ht!]
	\centering
	\begin{tabular}{ |p{3cm}||p{1.0cm}|p{1.0cm}|}
		\hline
		Predicted panel & 3 & 6 \\
		\hline
		Missing panel 3& 66 & 0   \\
		Missing panel 6& 0 & 66 \\
		\hline
	\end{tabular}
	\caption{Confusion Matrix of neural network classifier trained on the first dataset, test set, total accuracy: 100\%}
	\label{Tab:Second_set_conf_mat}
\end{table}

\section{Knowledge transfer between the two problems}
\label{sec:transfer}

Having split the problem into two sub-problems, a scheme motivated by transfer learning in deeper learners was examined. The idea being to establish if the features
extracted at the hidden layer in one problem, could be used for the other. In transfer learning termology, the seven-class problem specifies the source domain and 
task, while the two-class problem gives the target domain and task. The transfer is carried out by using the fixed input and hidden layers from the classifier in
the source task, as the input and hidden layers of the target task; this means that only the weights between the hidden and output layers remain to be trained for
the target task. This strategy reduces the number of parameters considerably. The functional form of the network for the source task is given by,

\begin{equation} 
\label{eq:network_output}
     \mathbf{y} = f_{0}W_2(f_{1}(W_{1}\mathbf{x} + b_{1}) + b_{2}
\end{equation}
where $f_0$ and $f_1$ are the non-linear activation functions of the output layer and the hidden layer respectively, $W_{1,2}$ are the weight matrices of the 
transformations between the layers, $b_{1, 2}$ are the bias vectors of the layers, $\mathbf{x}$ is the input vector and $\mathbf{y}$ the output vector. The {\em softmax} 
function is chosen to be the activation function of the decision layer, as this is appropriate to a classification problem. The prediction of the network, concerning 
which damage class the sample belongs to, is the index that maximises the output vector $\mathbf{y}$; the outputs are interpreted as the {\em a posteriori} probabilities
of class membership, so this leads to a Bayesian decision rule. Loosely speaking, one can think of the transformation between the hidden and ouput layers as the actual
classifier, and the transformation between the input layer into the hidden layer as a map to latent states in which the classes are more easily separable. In the context
of deep networks, the hope is that the earlier layers carry out an automated feature extraction which facilitates an eventual classifier. In the deep context, transfer
between problems is carried out by simply copying the `feature extraction' layers directly into the new network, and only training the later classificatiion layers. The
simple idea explored here, is whether that strategy helps in the much more shallow learner considered in this study. The transfer is accomplished by copying the weights $W_1$
and biases $b_1$ from sub-problem one directly into the network for sub-problem two, and only training the weights $W_2$ and biases $b_2$.

\begin{figure}[ht!]
 	\centering
 	\includegraphics[scale=0.70]{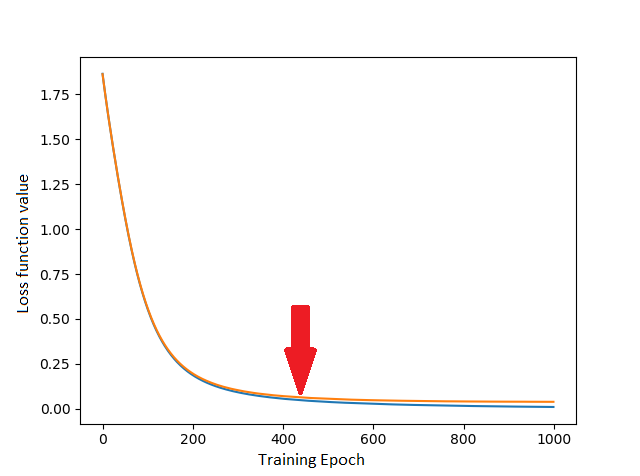}
 	\caption{Training and validation loss histories and the point of early stopping (red arrow).}
 	\label{fig:early_stoping}
\end{figure}

As before, multiple neural networks were trained on the first dataset. In a transfer learning scheme, it is even more important that models should not be overtrained, 
since that will make the model too case-specific and it would be unlikely for it to carry knowledge to other problems. To achieve this for the current problem, an early stopping strategy 
was followed. Models were trained until a point were the value of the loss function decreases less than a percentage of the current value. An example of this can be 
seen in Figure \ref{fig:early_stoping} where instead of training the neural network for 1000 epochs, training stops at the point indicated with the red arrow.

After multiple networks were trained following the early stopping scheme above, the network with the lowest value on validation loss was determined and the transfer 
learning scheme was applied to the second problem. The nonlinear transformation given by the transition from the input layer to the hidden layer was applied on the data 
of the second dataset. Consequently, another neural network was trained on the transformed data, having only one input layer and one output/decision layer. To comment 
on the effect of the transformation, another two-layer network was trained on the original second dataset and the results were compared. 

\begin{table}[ht!]
	\centering
	\begin{tabular}{ |p{3cm}||p{1.0cm}|p{1.0cm}|}
		\hline
		Predicted panel & 3 & 6 \\
		\hline
		Missing panel 3& 65 & 1   \\
		Missing panel 6& 2 & 64 \\
		\hline
	\end{tabular}
	\caption{Confusion Matrix of neural network classifier trained on the original data of the second dataset, test set, total accuracy: 97.72\%}
	\label{Tab:conf_original_data_2_layer}
\end{table}

\begin{table}[ht!]
	\centering
	\begin{tabular}{ |p{3cm}||p{1.0cm}|p{1.0cm}|}
		\hline
		Predicted panel & 3 & 6 \\
		\hline
		Missing panel 3& 65 & 1   \\
		Missing panel 6& 3 & 63 \\
		\hline
	\end{tabular}
	\caption{Confusion Matrix of neural network classifier trained on the transformed data of the second dataset, test set, total accuracy: 96.96\%}
	\label{Tab:conf_trans_data_2_layer}
\end{table}

\begin{figure}[ht!]
	\centering
	\begin{subfigure}{.5\textwidth}
		\centering
		\includegraphics[width=.8\linewidth]{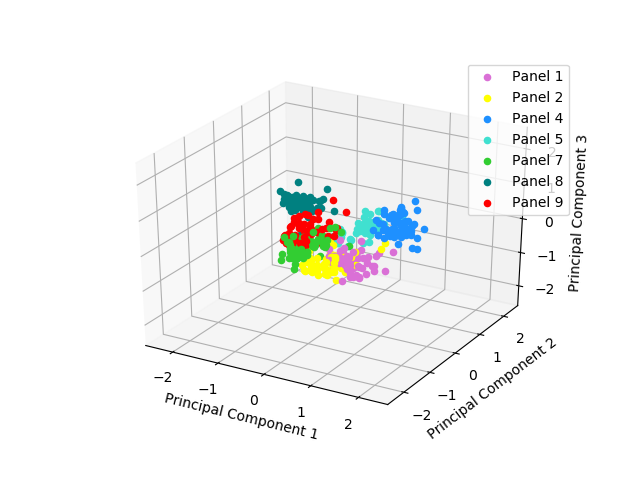}
		\caption{}
		\label{fig:first_dataset_features_pcs}
	\end{subfigure}%
	\begin{subfigure}{.5\textwidth}
		\centering
		\includegraphics[width=.8\linewidth]{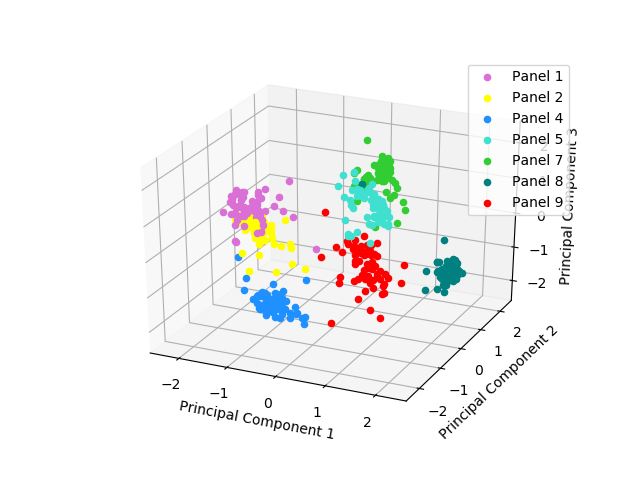}
		\caption{}
		\label{fig:first_dataset_features_pcs_transed}
	\end{subfigure}
	\caption{Principal components of original features of the first dataset (a) and transformed features (b).}
	\label{fig:first_pcs_1}	
\end{figure}

\begin{figure}[ht!]
	\centering
	\begin{subfigure}{.5\textwidth}
		\centering
		\includegraphics[width=.8\linewidth]{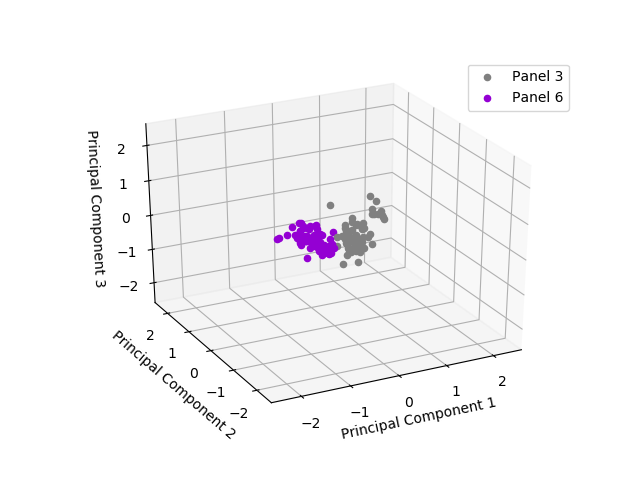}
		\caption{}
		\label{fig:second_dataset_features_pcs}
	\end{subfigure}%
	\begin{subfigure}{.5\textwidth}
		\centering
		\includegraphics[width=.8\linewidth]{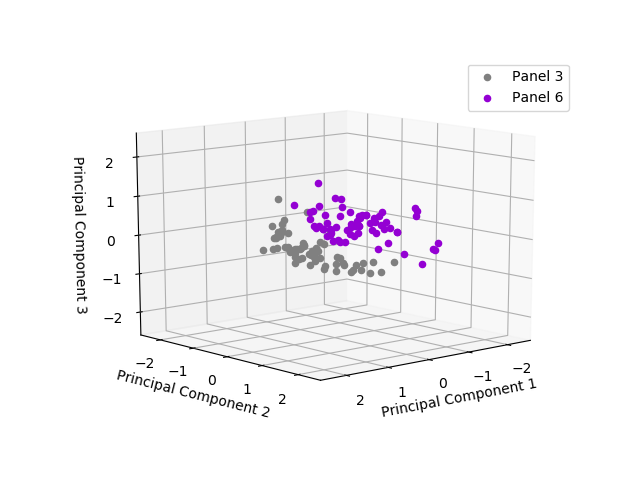}
		\caption{}
		\label{fig:second_dataset_trans_features_1_pcs}
	\end{subfigure}
	\caption{Principal components of original features of the second dataset (a) and transformed features (b).}
	\label{fig:second_pcs}	
\end{figure}

The confusion matrices of the two neural networks on the testing data are given in Tables \ref{Tab:conf_original_data_2_layer} and \ref{Tab:conf_trans_data_2_layer}; 
the misclassification rates are very similar. However, it is interesting to also look at the effect of the transfer on the convergence rate of the network trained 
on the transferred data and also to illustrate the feature transformation on the first and the second datasets.

\begin{figure}[ht!]
	\centering
	\includegraphics[scale=0.5]{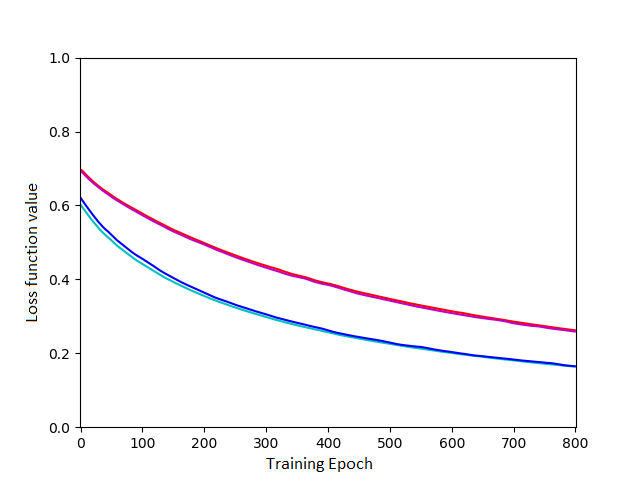}
	\caption{Loss histories of transferred model: train(blue), validation(cyan) and model trained on initial data: train(red), validation(magenta).}
	\label{fig:loss_histories}
\end{figure}

The training histories of the two models can be seen in Figure \ref{fig:loss_histories}. It is clear that the loss history of the model with transformed data (blue 
and cyan lines) converges faster, especially in the initial part of the training, and it also reaches a lower minimum value for the loss function in the same number 
of training epochs. This can be explained by looking at the effect of the learnt transformation on the data. In Figures \ref{fig:second_pcs} and \ref{fig:first_pcs_1} 
this effect is illustrated. (Note that the points are different from those in Figure \ref{fig:first_pcs}, because principal component analysis was performed this time 
on the normalised data in the interval [-1, 1] for the neural network training). The transformation spreads out the points of the original problem (first dataset) in 
order to make their separation by the decision layer easier; however, it is clear that it also accomplishes the same result on the second dataset. The points in Figure \ref{fig:second_dataset_trans_features_1_pcs} are spread out compared to the initial points and thus, their separation by the single layer neural network is easier. 
Furthermore, the points lay further away from the required decision boundary and this explains both the faster training convergence and the lower minimum achieved. 
In contrast to the transformation of the first dataset, in the second dataset, the transformation does not concentrate points of the same class in specific areas of the feature space. 
In Figure \ref{fig:first_dataset_features_pcs_transed} the points are both spread and concentrated closer according to the class they belong. This probably means that 
only a part of the physics of the problem is transferred in the second problem through this specific transformation.

\section{Discussion and Conclusions}
\label{sec:conclusions}

For the SHM classification (location) problem considered here, splitting the dataset into two subsets contributed to increasing the classification accuracy by a 
small percentage. This result was explained by the lesser effect that the small panel removals had on the novelty index features. This issue arose because the 
points representing these classes were close to each other and also points from other classes -- those corresponding to large panel removal/damage. By considering 
the two damage cases as different problem, perfect accuracy was achieved in the task of classifying damage to the small panels, and there was also a small increase 
in the performance of the classifier tasked to identify the more severe damage states.

An attempt at a crude form of transfer learning was also investigated. Having trained the neural network classifier on the first dataset of the seven damage cases, 
transfer of knowledge to the second sub-problem was considered. This was accomplished by copying the first two layers of the first classifier -- the `feature 
extraction' layers -- directly into the second classifier and only training the connections from the hidden layer to the output. The result is not particularly 
profound; the transfer does allow a good classifier, even with the smaller set of trainable parameters, but is not as good as training the network from scratch.
The result is interesting, because it is clear that the source network is carrying out a feature clustering and cluster separation on the source data, that is still
useful when the target data are presented. This suggests that the main issue with the small-panel damage classification is that the data are masked by the close 
presence of the large-panel data. Separating out the small-panel is the obvious answer. The results are interesting because they illustrate in a `toy' example, how
the early layers in deeper networks are manipulating features automatically in order to improve the ultimate classification step. The other benefit of the 
separation into sub-problems, was the faster convergence of the network training.

\section{Acknowledgement}
\label{sec:ack}

The authors would like to acknowledge Graeme Manson for providing the data used. Moreover, the authors would like to acknowledge the support of the Engineering and Physical Science Research Council (EPSRC) and the European Union (EU). G.T is supported by funding from the EU’s Horizon 2020 research and innovation programme under the Marie Skłodowska-Curie grant agreement DyVirt (764547). The other authors are supported by EPSRC grants EP/R006768/1, EP/R003645/1, EP/R004900/1 and EP/N010884/1.

% References.
\bibliographystyle{unsrt}
\bibliography{imac_20_GT_2}

\end{document}